\pdfoutput=1

\documentclass[11pt]{article}

\usepackage[]{acl}

\usepackage{times}
\usepackage{latexsym}
\usepackage[colorinlistoftodos]{todonotes}
\usepackage{longtable}
\usepackage{tabularx}
\usepackage{multirow}
\usepackage[normalem]{ulem}
\usepackage{fancyvrb}
\usepackage{makecell}
\usepackage{xcolor,colortbl}
\usepackage{comment}
\usepackage{subcaption}

\usepackage[T1]{fontenc}

\usepackage[utf8]{inputenc}

\usepackage{microtype}

\newcommand{\red}[1]{\textcolor{red}{#1}}
\newcommand{\wip}[1]{\textcolor{black}{#1}}

\title{Human Evaluation and Correlation with Automatic Metrics \\ in Consultation Note Generation}

\author{
Francesco Moramarco\textsuperscript{†}\textsuperscript{‡},
Alex Papadopoulos Korfiatis\textsuperscript{†},
Mark Perera\textsuperscript{†},
Damir Juric\textsuperscript{†},
\\
\bf{Jack Flann\textsuperscript{†}},
\bf{Ehud Reiter\textsuperscript{‡}},
\bf{Anya Belz\textsuperscript{‡}},
\bf{Aleksandar Savkov\textsuperscript{†}}
\\
\textsuperscript{†}Babylon
\textsuperscript{‡}University of Aberdeen
\\
\textsuperscript{†} \texttt{\{francesco.moramarco, alex.papadopoulos,
mark.perera,}
\\
\texttt{damir.juric, jack.flann, sasho.savkov\}@babylonhealth.co.uk}
\\
\textsuperscript{‡} \texttt{\{r01fm20, ehud.reiter, anya.belz\}@abdn.ac.uk}}

\begin{document}
\maketitle
\begin{abstract}

In recent years, machine learning models have rapidly become better at generating clinical consultation notes; yet, there is little work on how to properly evaluate the generated consultation notes to understand the impact they may have on both the clinician using them and the patient's clinical safety.

To address this we present an extensive human evaluation study of consultation notes where 5 clinicians (i) listen to 57 mock consultations, (ii) write their own notes, (iii) post-edit a number of automatically generated notes, and (iv) extract all the errors, both quantitative and qualitative. We then carry out a correlation study with 18 automatic quality metrics and the human judgements. We find that a simple, character-based Levenshtein distance metric performs on par if not better than common model-based metrics like BertScore. All our findings and annotations are open-sourced.

\end{abstract}
\section{Introduction}
\label{sec:intro}

Modern Electronic Health Records (EHR) systems require clinicians to keep a thorough record of every patient interaction and management decision. While this creates valuable data that may lead to better health decisions, it also significantly increases the burden on the clinicians, with studies showing this is a major contributor to burnout \citep{arndt2017tethered}.

In most primary healthcare practices, the universal record of a clinician-patient interaction is the SOAP (Subjective, Objective, Assessment, Plan) note, which captures the patient's history, and the clinician's observations, diagnosis, and management plan \citep{pearce2016essential}. At the end of a consultation, the clinician is required to write up a SOAP note of the encounter. With the exception of the clinician's internal observations on how the patient looks and feels, most of the SOAP note is verbalised and could be automatically constructed from the transcript of the consultation.

A number of recent studies \cite{enarvi2020generating,joshi2020dr,zhang2021leveraging} propose using summarisation systems to automatically generate consultation notes from the verbatim transcript of the consultation---a task henceforth referred to as Note Generation. Yet, there is very limited work on how to evaluate a Note Generation system so that it may be safely used in the clinical setting. Where evaluations are present, they are most often carried out with automatic metrics; while quick and cheap, these metrics were devised for general purpose summarisation or machine translation, and it is unclear whether they work just as well on this new task. 
In the field of automatic summarisation and Natural Language Generation (NLG) in general, human evaluation is the gold standard protocol.
Even in cases where the cost of using human evaluation is prohibitive, it is essential to establish the ground truth scores which automatic metrics should aim for.

Our contributions are: (i) a large-scale human evaluation performed by 5 clinicians on a set of 285 consultation notes, (ii) a thorough analysis of the clinician annotations, and (iii) a correlation study with 18 automatic metrics, discussing limitations and identifying the most suitable metrics to this task. We release all annotations, human judgements, and metric scores.\footnote{\url{https://github.com/babylonhealth/primock57}}


\renewcommand{\cellalign}{l}

\begin{table*}[t]
    \setlength{\tabcolsep}{4pt} 
    \def\arraystretch{1.3}

    \centering
    \begin{tabular}{l|p{7cm}|p{7cm}}
        \multicolumn{2}{c|}{\cellcolor{blue!25}\textbf{Transcript}} & \multicolumn{1}{c}{\cellcolor{blue!25}\textbf{Note}} \\\hline
          \cellcolor{blue!5}Clinician & \cellcolor{blue!5}Hello. & \multirow{8}{*}{\makecell{3/7 hx of diarrhea, mainly watery. \\ 
 No blood in stool. Opening bowels x6/day.\\
 Associated LLQ pain - crampy, intermittent, \\ nil radiation.\\
 Also vomiting - mainly bilous.\\
 No blood in vomit.\\
 Fever on first day, nil since.\\
 Has been feeling lethargic and weak since.\\
 Takeaway 4/7 ago - Chinese restaurant.\\
 Wife and children also unwell with vomiting, \\ but no diarrhea. No other unwell contacts.\\
 PMH: Asthma\\
 DH: Inhalers\\
 SH: works as an accountant.\\
 Lives with wife and children.\\
 Affecting his ADLs as has to be near toilet.\\
 Nil smoking/etOH hx}}\\\cline{1-2}
         
         Patient & Hello, how are you? \\\cline{1-2}
         \cellcolor{blue!5}Clinician & \cellcolor{blue!5}Hello. How can I help you this morning? \\\cline{1-2}
         Patient & All right. I just had some diarrhea for the last three days and it's been affecting me. I need to stay close to the toilet. And yeah, it's been affecting my day-to-day activities. \\\cline{1-2}
         \cellcolor{blue!5}Clinician & \cellcolor{blue!5}I'm sorry to hear that and when you say diarrhea, what do you mean by diarrhea? Do you mean you're going to the toilet more often or are your stools more loose? \\\cline{1-2}
         Patient & \makecell{Yeah, so it's like loose and watery \underline{\textbf{stole}}\\going to the toilet quite often.}\\\cline{1-2}
         \cellcolor{blue!5}Clinician & \cellcolor{blue!5}\underline{\textbf{freak}} \\
         & ... &\\
    \end{tabular}
    \caption{Snippet of a mock consultation transcript and the Subjective part of the corresponding SOAP note. The transcript is produced by Google Speech-to-text\textsuperscript{3}; the bold-underlined text shows transcription errors. The note is written by the consulting clinician.}
    \label{tab:transcript_note}
\end{table*}

\begin{figure*}[t]
    \centering
    \includegraphics[width=1.0\textwidth]{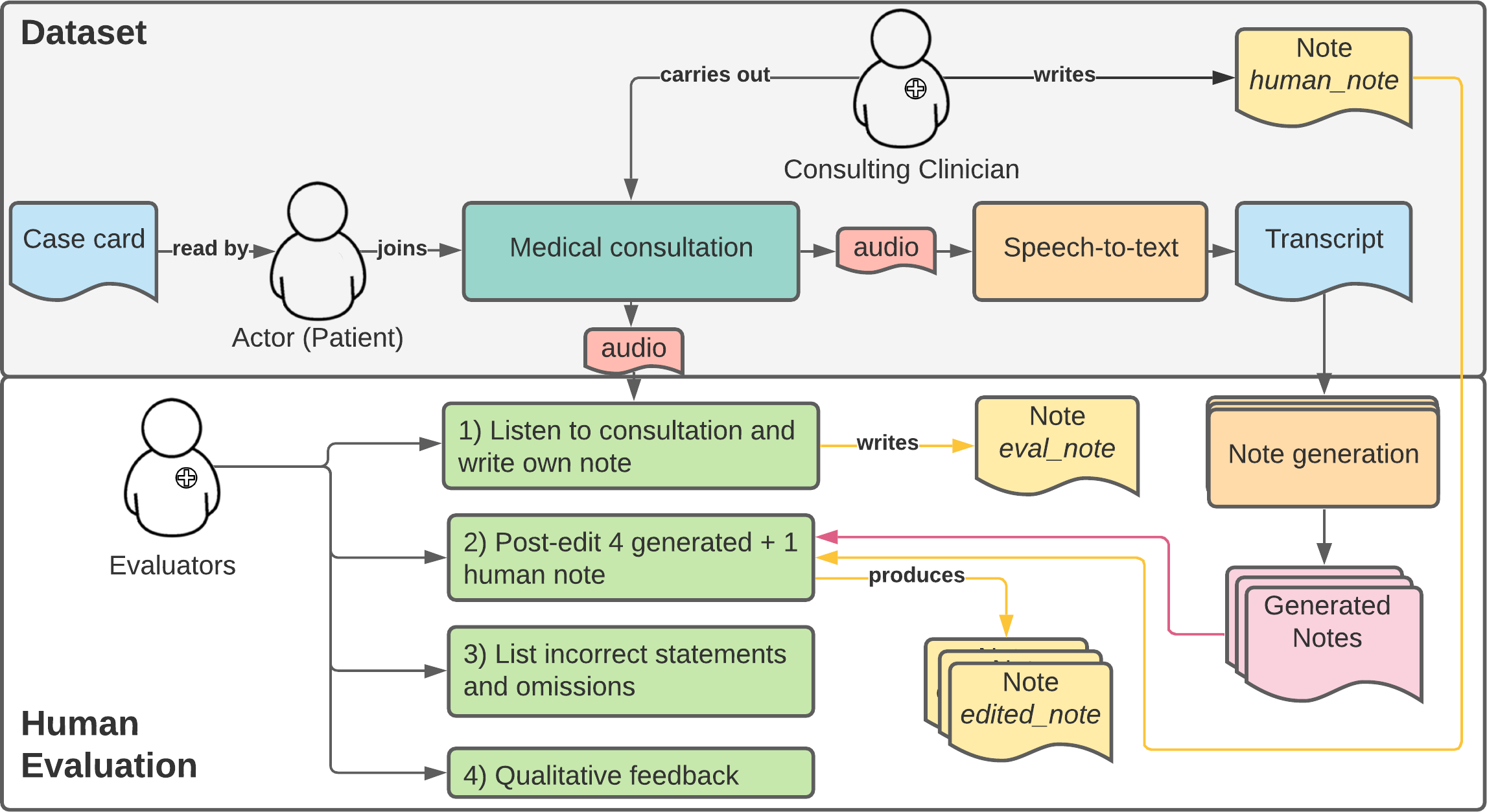}
    \caption{Diagram of the dataset creation and the four tasks involved in the human evaluation.}
    \label{fig:diagram}
\end{figure*}

\begin{figure*}[t]
    \centering
    \includegraphics[width=1.0\textwidth]{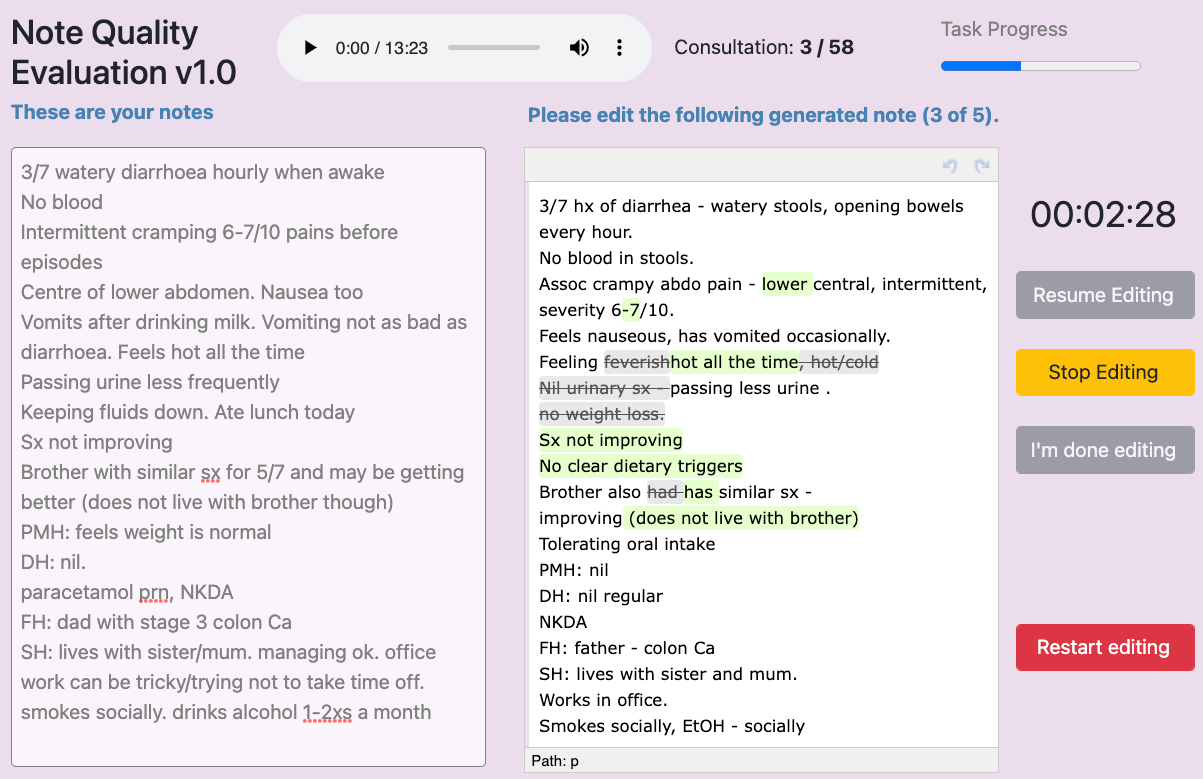}
    \caption{Screenshot of the post-editing task where the evaluator is correcting a note with the track-changes interface. Text in green shows what they have added, and text in grey (strikethrough) what they have deleted.}
    \label{fig:edit-note}
\end{figure*}

\begin{figure}[t]
    \centering
    \includegraphics[width=.5\textwidth]{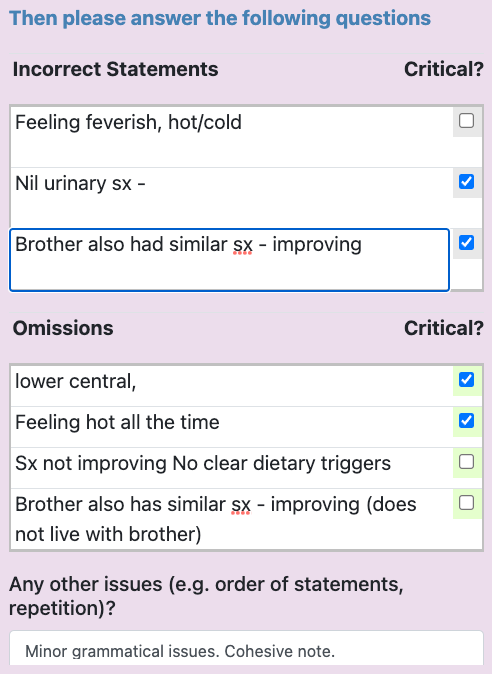}
    \caption{Screenshot of the scoring task, where the clinician is asked to quantify the incorrect statements and omissions in the generated note.}
    \label{fig:score-note}
\end{figure}
\section{Related Work}
\label{sec:related-work}

Note Generation has been in the focus of the academic community with both extractive methods \cite{moen2016comparison, alsentzer2018extractive}, and with abstractive neural methods \cite{zhang2018learning, liu2019topic, macavaney2019ontology, zhang2020optimizing, enarvi2020generating, joshi2020dr,krishna-etal-2021-generating, chintagunta2021medically,yim2021towards, moramarco2021preliminary, zhang2021leveraging}.
Whether these studies discuss the generation of radiology reports, patient-nurse summaries, discharge summaries, or SOAP notes, they all deal with long passages of text in the medical domain. This is a critical distinction from other application contexts (e.g. news summarisation): here, commonly used and well-studied evaluation criteria such as `fluency', `relevance', and `adequacy' are superseded by other criteria, such as `omissions of important negatives', `misleading information', `contradictions', etc. 
\wip{In addition, common summarisation metrics such as ROUGE \cite{lin2004rouge} or BertScore \cite{zhang2019bertscore} measure the standalone quality of outputs and are not typically evaluated against more extrinsic criteria, such as post-editing times.}
Of the 18 studies on the subject that we could identify, 13 present an automatic evaluation (typically based on ROUGE and sometimes on medical entity linking) and 12 carry out a small-scale intrinsic human evaluation. 
In particular, \citet{moen2016evaluation} employ three domain experts to review 40 generated notes with Likert scales along 30 criteria (including `Long-term diagnosis', `Reason for admission', `assessment'), but report that the subjects found the 30 item scale too difficult and detailed to assess. \citet{macavaney2019ontology} use one domain expert to review 100 notes and report Likert scale values for `Readability', `Accuracy', and `Completeness'. \citet{moramarco2021preliminary} employ three clinicians and compare the times to post-edit generated notes with those of writing them from scratch, reporting that, while faster, post-editing may be more cognitively intensive than writing.  

Outside of the medical domain, our work is comparable to \citet{fabbri2021summeval}, who run an automatic metrics correlation study for news article summaries for the CNN/DailyMail dataset \cite{nallapati2016abstractive}. They also release code\footnote{\url{https://github.com/Yale-LILY /SummEval}} for evaluating text with a suite of common metrics, some of which we include in our own list of metrics to evaluate.
\section{Dataset and Models}

Our evaluation study is based on a dataset of 57 pairs of mock consultation transcripts and summary notes \citep{primock2022}.\footnote{The dataset is available at: \url{https://github.com/babylonhealth/primock57}}
The data was produced by enacting consultations using clinical case cards. The clinicians that conducted the mock consultations also wrote the corresponding SOAP note. The consultations span common topics within primary healthcare and are about 10 minutes long.

To mimic a live clinical environment, the audio of the consultations was transcribed with Google Speech-to-text engine\footnote{\label{google-stt}\url{https://cloud.google.com/speech-to-text}}. These transcripts form the input to the Note Generation models. The aim is to generate the Subjective part of a SOAP note. Table \ref{tab:transcript_note} shows an example transcript and respective note. Figure \ref{fig:diagram} describes the creation of the dataset and how the data feeds into the human evaluation tasks described below.


In a fashion similar to \citet{chintagunta2021medically, moramarco2021preliminary, zhang2021leveraging}, we fine-tune 10 neural summarisation models based on BART \cite{lewis-etal-2020-bart} on a proprietary dataset of 130,000 real consultation notes and transcripts. In accordance with our evaluation dataset, the training set consists of automatic Google Speech-to-text transcripts as inputs and the Subjective part of the corresponding notes as outputs.

The base models are large BART architectures pretrained on the CNN/Dailymail dataset\footnote{\url{https://huggingface.co/facebook/bart-large-cnn}}.
Since our focus is on evaluation, the aim was to obtain models which would produce different outputs to cover a wider range of errors. The differences between the models included: 
fine-tuning on different sized datasets; using pre-processing techniques such as filtering the transcripts for relevant sentences; and using post-processing techniques such as filtering the generated notes for irrelevant sentences.
\section{Human Evaluation Setup}
\label{sec:human-eval-setup}

Under the supervision of one of the authors (a clinician expert in AI development henceforth referred to as the Lead Clinician) we design the following evaluation tasks:

\begin{enumerate}
    \item \textbf{Listen to the mock consultation audio and take notes (\emph{eval\_notes}).} These \emph{eval\_notes} appear on the evaluator screen throughout to help reduce the cognitive load of remembering what was discussed in the consultation. 

    \item \textbf{Relying on the \emph{eval\_notes} and the consultation audio, read 5 different notes and post-edit each one of them.} Post-editing \wip{consists of correcting an imperfect note to produce a factually accurate and relevant note} \cite{sripada2005evaluation}. It mimics how a synthetic note could be used in clinical practice while also bootstrapping the error identification \cite{moramarco2021preliminary}. For this purpose, the evaluation platform includes a track-changes interface, which highlights insertions and deletions (Figure \ref{fig:edit-note}), and records the time taken to post-edit.
    
    \item \textbf{For each note, classify the errors into two categories: `incorrect statements' and `omissions', by copying the spans of text from the post-editing interface and pasting them in the appropriate table (as in Figure \ref{fig:score-note}).} We define `incorrect statements' as sentences in the generated notes which contain one or more factual errors (compared to the consultation audio). Conversely, `omissions' are medical facts which should be recorded in a consultation note and were omitted by the model. Examples and edge cases (which were given to the evaluators for training) can be found in the Appendix, Figure \ref{fig:evaluators-instructions}.
    Each error is also tagged as `critical' if the information contained has essential clinical importance. Specifically, if the error would lead to medico-legal liability.
    
    \item \textbf{Report any qualitative feedback (e.g. regarding order of statements, repetition) in the `Other issues' box.} Figure \ref{fig:diagram} (bottom half) shows a diagram of the human evaluation workflow.
\end{enumerate}

The subjects of the study were 5 regularly practising clinicians (GPs) with a minimum of 3 years experience. As part of our ethical consideration, all clinicians were paid the UK standard GP working rate and were free to cease participation at any time if they wished. For diversity and inclusion, 2 male clinicians and 3 female clinicians were enlisted from a range of ethnic backgrounds. 

Following the tasks described above, each clinician evaluated the entire dataset of 57 mock consultations. Each consultation included 5 notes to evaluate, 4 of which were sampled from our 10 models and 1 was written by the consulting doctor (\emph{human\_note}). We shuffled these for every consultation and---to avoid biases---did not specify that one of the notes was not synthetic. 

The evaluation study took circa 30 working hours per evaluator to complete over a period of 8 weeks. Before commencing, each evaluator went through a training and practice process conducted by the Lead Clinician, who explained the evaluation interface and guided them through the annotation of a practice note. A copy of the evaluator instructions can be found in Appendix \ref{app:instructions}. Throughout the study, the authors and the Lead Clinician held weekly sessions with each evaluator where we shadowed the evaluation tasks through screen sharing. This helped us understand the difficulties in performing the tasks while ensuring the evaluators followed the guidelines set out for them.

\begin{table}[t]
    \centering
    \begin{tabular}{l|l|l}
        \textbf{Criterion} & \textbf{Agree.} & \textbf{Word Overlap}\\\hline
        Post-edit times & 0.542\textsuperscript{*} & --  \\
        Incorrect statements & 0.541\textsuperscript{*} & 0.431\textsuperscript{†}   \\
        Omissions & 0.374\textsuperscript{*} & 0.268\textsuperscript{†}  \\

    \end{tabular}
    \caption{Inter Annotator Agreement. \textsuperscript{*}Krippendorff's Alpha. \textsuperscript{†}Word-level F1 score.} 
    \label{tab:agreement}
\end{table}

\section{Results analysis}

\subsection{Agreement}
The result of the human evaluation consists of 285 evaluator notes (57 consultations x 5 evaluators), 1,425 post-edited notes (285 x 5 notes per consultation), post-editing times, count and spans of incorrect statements, count and spans of omissions, whether they are critical, and qualitative comments. When compared with more common evaluation approaches such as Likert scales and ranking methods, we believe our set-up provides a more granular and more interpretable set of judgements, albeit at the price of lowering the inter-annotator agreement. To compensate for this, the 5 evaluators annotate the same 57 tasks \cite{sheng2008get} and the scores are averaged in the correlation study (see Section \ref{sec:correlation-study}).



As shown in Table \ref{tab:agreement}, we compute inter-annotator agreement on the post-editing times, incorrect statements, and omissions. The absolute post-editing times are converted to a list of rankings for each evaluator, and agreement is computed with Krippendorff's alpha \cite{krippendorff2018content} with `ordinal' level of measurement. This ensures only the ranking of each note is captured in the agreement and not the editing speed of each evaluator. For example, where evaluator 1 takes 60 seconds and 120 seconds to post-edit two given notes and evaluator 2 takes 180 seconds and 240 seconds respectively, their agreement would be perfect because they both agreed that note 1 is quicker to edit than note 2.
Conversely, for incorrect statements and omissions we calculate `interval' Krippendorff's Alpha on the counts of errors identified by the evaluators. As the counts don't ensure that two evaluators have selected the same statements, we also compute word overlap F1 score as suggested by \citet{popovic2021reproduction}. As shown in Table \ref{tab:agreement}, the agreements for times and incorrect statements are not very strong (\citet{krippendorff2018content} indicate that $\alpha \geq 0.667$ is the lowest conceivable limit). We investigate the source of disagreement and attribute it to two main factors: (i) human error due to the difficulty inherent in the task, and (ii) stylistic differences in note writing. Examples of human error can be found in subsection \ref{sec:human-error}. As for stylistic differences, these are especially evident in the Omissions category, where some clinicians are thorough in their note taking and others only document the most important facts. See Appendix \ref{app:pairwise-agreement} for more details on pairwise agreement.



\subsection{Human error}
\label{sec:human-error}

\begin{table}[t]
    \centering
    \begin{tabular}{l|l|l}
      \textbf{Criterion} & \textbf{Human} & \textbf{Generated} \\\hline
     \textbf{Post-edit times}             & 96.5s & 136.4s \\
     \textbf{Number of Incorrect} & 1.3 & 3.9 \\
     \textbf{Number of Omissions} & 3.9 & 6.6 \\\hline
     \textbf{Note length}          &  16.9 & 21.5 \\
\end{tabular}
    \caption{Aggregated judgements for \emph{human} notes and generated notes. Note length is in number of sentences.}
    \label{tab:human-vs-generated}
\end{table}

To compare the accuracy of the models against human-written notes, we average all the judgements \wip{for our criteria} (post-edit times, incorrect statements, and omissions), \wip{aggregate by} the generated notes and the \emph{human\_notes} respectively, and report the results in Table \ref{tab:human-vs-generated}.
As expected, the \emph{human\_notes} performed better for all criteria; in particular, they contain fewer omissions while being on average 4.6 sentences shorter. However, the evaluators found imperfections in human notes too: it takes over 1.5 minutes on average to read and post-edit a \emph{human\_note}, and it contains over 1 incorrect statement and almost 4 omissions on average. While the omissions can be reconciled as stylistic differences among evaluators, the incorrect statements are potentially more impactful. To investigate, we select two human notes and ask the Lead Clinician to post-edit them, comparing the results with those of the evaluators. In the first case, the Lead Clinician agrees with the evaluators in that the human note contains the following two incorrect statements:

\begin{flushleft}
\setlength\tabcolsep{4pt} 
\begin{tabular}{p{3.0cm}|p{4cm}}
     \textbf{Inc. statement} & \textbf{Correction} \\\hline
     Also vomiting -- mainly bilous & Also vomiting\\\hline
     Wife and children also unwell with vomiting, but no diarrhea. & One child had some vomiting, but no other symptoms in wife and other child. \\
\end{tabular}
\end{flushleft}

Upon inspecting the consultation recording, the Lead Clinician found that the word `bilious' was not stated by the patient. However, the consulting clinician may have used this term due to a personal habitual documentation style (as clinically, vomit with no red flags can conventionally be referred to as bilious). The words `Wife and children also unwell with vomiting, but no diarrhea' were not stated by the patient. Instead, the patient made a tangential statement summarised here: `One child had some vomiting but no other symptoms in wife and other child.' Therefore, it is inferred that this clinician likely made a normal human error due to excessive patient detail (non-critical).

In the second case, the Lead Clinician found no issues with the human note. Upon inspecting the corrections from the evaluators, he concluded that what they selected as incorrect statements were medical conditions inferred by the consulting clinician yet not specifically stated by the patient. 
We highlight this to show that it is unclear whether the task has a single ground truth, as even human experts don't completely agree; well thought-out evaluation tasks can mitigate this and produce one or more good ground truth approximations. Detailed examples can be found in Appendix \ref{app:human-errors}.

\subsection{Analysis of criteria}
\label{sec:analysis-of-criteria}

To understand the interdependence between our criteria, we compute Pearson's correlation \cite{freedman2007statistics} and Spearman's rank correlation \cite{zar2005spearman} coefficients between each pair. 

Table \ref{tab:criteria-correlation} shows a moderately strong correlation between the time it takes to post-edit a note and the number of incorrect statements it contains. The correlation between post-edit times and omissions is stronger, which could be explained by the fact that it takes longer to type an omitted statement than to delete or edit an incorrect one. Finally, the correlation between post-edit times and \textit{incorrect+omissions} is strong, which suggests that post-edit times is a function of the number of edits and that one of these criteria could be a proxy for the other.

We also compute the correlation between each criterion and the length of the generated note. These numbers can be used as a benchmark for automatic metric correlation; for example, the 0.413 Spearman's correlation between post-edit times and note length indicates that any automatic metric needs to surpass this value in order to be more useful than simply counting the number of sentences in the note.

\begin{table}[t]
    \setlength{\tabcolsep}{4pt} 
    \centering
    \begin{tabular}{l|l|c|c}
\rowcolor[HTML]{FFFFFF} 
\textbf{Criterion 1} & \textbf{Criterion 2} & \textbf{Pears.}              & \textbf{Spear.}             \\\hline
\rowcolor[HTML]{FFFFFF} 
Post-edit times         & Incorrect       & \cellcolor[HTML]{91D0A2}0.543 & \cellcolor[HTML]{86CB98}0.599 \\
\rowcolor[HTML]{FFFFFF} 
Post-edit times         & Omissions       & \cellcolor[HTML]{63BD7B}0.769 & \cellcolor[HTML]{61BC79}0.781 \\
\rowcolor[HTML]{FFFFFF} 
Post-edit times         & Inc + Omi       & \cellcolor[HTML]{5DBA76}0.8   & \cellcolor[HTML]{57B771}0.829 \\
\rowcolor[HTML]{FFFFFF} 
Post-edit times         & Note length     & \cellcolor[HTML]{B2DEBE}0.38  & \cellcolor[HTML]{ACDCB8}0.413 \\
\rowcolor[HTML]{FFFFFF} 
Incorrect       & Omissions       & \cellcolor[HTML]{B2DEBD}0.384 & \cellcolor[HTML]{A1D7AF}0.467 \\
\rowcolor[HTML]{FFFFFF} 
Incorrect       & Note length     & \cellcolor[HTML]{92D1A3}0.537 & \cellcolor[HTML]{96D2A6}0.52  \\
\rowcolor[HTML]{FFFFFF} 
Omissions       & Note length     & \cellcolor[HTML]{E7F5EB}0.122 & \cellcolor[HTML]{DAF0E0}0.183

    \end{tabular}
    \caption{Correlation coefficients between the criteria; all numbers statistically significant (p value < 0.001). A darker shade of green means higher correlation.}
    \label{tab:criteria-correlation}
\end{table}

\subsection{Qualitative results}

\begin{table*}[t]
    \centering
    \begin{tabular}{p{3.3cm}|p{10.6cm}|c}
        \textbf{Issue} & \textbf{Examples} & \textbf{Occ.} \\\hline
        \multicolumn{3}{l}{\cellcolor{blue!25}\small \textbf{Discourse level}} \\ \hline
        Contradiction & \emph{no family history of bowel issues. \newline father has history of colon cancer} & 25\\\hline
        Contradiction not reported & patient corrected herself but note did not pick it up. & 4\\\hline
        Symptom mentioned is reported as fact &  \emph{tingling of hands} stated by clinician (not refuted/confirmed by pt) & 18\\\hline
        Misleading statement & statement: \emph{not working} when patient has been off ill for a few days due to current sickness reads like patient is unemployed & 9\\\hline
        
        \multicolumn{3}{l}{\cellcolor{blue!25}\small \textbf{Factual errors}} \\\hline
        Hallucination & \emph{at home and in a private place} was not mentioned in the consultation & 17\\\hline
        Incorrect statement & statement: \emph{brother has diabetes}. Correction: \emph{mother has diabetes} & 34\\\hline
        Nonsensical & \emph{No recent unwell with diarrhoea} & 18 \\\hline
        
        \multicolumn{3}{l}{\cellcolor{blue!25}\small \textbf{Stylistic errors}} \\ \hline
        Repetition & \emph{loose and watery stools \newline stool is mainly watery} & 93\\\hline
        Incorrect order of statements & \emph{heart attack} should be in PMH; structure of history a bit jumbled; recorded social smoker in alcohol section. & 38\\\hline
        Use of not universally recognised acronyms & NRS/EMS/DOA are not standard acronyms &7\\\hline
        
        \multicolumn{3}{l}{\cellcolor{blue!25}\small \textbf{Omissions}} \\ \hline
        Generic & no mention of \emph{unable to open bowels} & 57\\\hline
        Omissions of important negatives & \emph{No fever \newline No shortness of breath} & 5\\\hline
        
        \multicolumn{3}{l}{\cellcolor{blue!25}\small \textbf{Other}}\\\hline
        Good behaviour & Contains all the history that was covered in the audio and follows a logical structure. & 21\\\hline
    \end{tabular}
    \caption{Taxonomy of errors gathered through the qualitative feedback from the evaluators.}
    \label{tab:other-issues}
\end{table*}

As introduced in Section \ref{sec:human-eval-setup}, the evaluators provide qualitative feedback about the generated notes in the `Other Issues' field. When analysing these comments, a number of repeated patterns emerged, highlighting common pitfalls in the generated notes. Based on these we defined a small taxonomy (Table \ref{tab:other-issues}; issues in the \emph{human\_notes} are excluded), providing examples and occurrences of each issue type. Aside from incorrect statements and omissions, the most significant issues revolve around repetition, disjointed notes, and contradiction. Upon investigating, we believe that all three are related to the tendencies of the models to generate the consultation note following the chronological order of the transcript. While that is an intuitive behaviour, consultations are seldom carried out in the order of SOAP note sections (Subjective, Objective, Assessment, Plan), with the patient providing relevant information whenever they can, sometimes after the clinician has discussed assessment and plan.

\section{Correlation with Automatic Metrics}
\label{sec:correlation-study}

\begin{table*}[t]
    \setlength{\tabcolsep}{3.2pt} 
    \centering
    \def\arraystretch{.9} 
    \begin{tabular}{l|ccc|c|c|c|c|c|c|c|c}
         \textbf{Criterion:} & \multicolumn{5}{c|}{\textbf{Post-edit times}} & \multicolumn{2}{c|}{\textbf{Inc+Omi}} & \multicolumn{2}{c|}{\textbf{Incorrect}} & \multicolumn{2}{c}{\textbf{Omissions}}\\\hline
         \textbf{Reference:} & \textbf{human}  & \textbf{edited} & \textbf{eval} & \textbf{avg} & \textbf{max} & \textbf{avg} & \textbf{max} & \textbf{avg} & \textbf{max} & \textbf{avg} & \textbf{max}\\\hline
    
\textbf{ROUGE-1-F1\textsuperscript{*}}       & \cellcolor[HTML]{BCE2C6}0.334          & \cellcolor[HTML]{80C994}0.627          & \cellcolor[HTML]{DFF2E4}0.160          & \cellcolor[HTML]{A6D9B3}0.443          & \cellcolor[HTML]{90D0A1}0.550          & \cellcolor[HTML]{8ACD9C}0.580          & \cellcolor[HTML]{71C286}0.704          & \cellcolor[HTML]{B3DFBE}0.378          & \cellcolor[HTML]{99D4A9}0.505          & \cellcolor[HTML]{8ECF9F}0.561          & \cellcolor[HTML]{7BC790}0.651          \\
\textbf{ROUGE-2-F1\textsuperscript{*}}       & \cellcolor[HTML]{B2DEBD}0.384          & \cellcolor[HTML]{7BC78F}0.653          & \cellcolor[HTML]{DEF1E3}0.166          & \cellcolor[HTML]{90D0A1}0.551          & \cellcolor[HTML]{8CCE9D}0.570          & \cellcolor[HTML]{73C388}0.694          & \cellcolor[HTML]{6BC082}\textbf{0.731} & \cellcolor[HTML]{9AD4A9}0.501          & \cellcolor[HTML]{8ECFA0}0.557          & \cellcolor[HTML]{7DC891}0.641          & \cellcolor[HTML]{7BC78F}0.654          \\
\textbf{ROUGE-3-F1\textsuperscript{*}}       & \cellcolor[HTML]{B5E0C1}0.366          & \cellcolor[HTML]{7DC791}0.645          & \cellcolor[HTML]{E8F5EB}0.117          & \cellcolor[HTML]{8BCD9C}\textbf{0.576} & \cellcolor[HTML]{8DCE9E}0.565          & \cellcolor[HTML]{6AC081}\textbf{0.734} & \cellcolor[HTML]{6BC082}\textbf{0.731} & \cellcolor[HTML]{8FCFA0}0.555          & \cellcolor[HTML]{8CCE9E}\textbf{0.568} & \cellcolor[HTML]{79C68D}0.663          & \cellcolor[HTML]{7CC790}0.646          \\
\textbf{ROUGE-4-F1\textsuperscript{*}}       & \cellcolor[HTML]{BAE2C5}0.342          & \cellcolor[HTML]{7FC993}0.632          & \cellcolor[HTML]{F0F9F2}0.076          & \cellcolor[HTML]{8BCD9D}0.575          & \cellcolor[HTML]{8ECFA0}0.557          & \cellcolor[HTML]{68BF7F}\textbf{0.745} & \cellcolor[HTML]{6CC083}0.726          & \cellcolor[HTML]{8ACD9C}\textbf{0.581} & \cellcolor[HTML]{8BCE9D}\textbf{0.573} & \cellcolor[HTML]{79C68E}0.661          & \cellcolor[HTML]{7EC892}0.636          \\
\textbf{ROUGE-L-Pr\textsuperscript{*}}       & \cellcolor[HTML]{B9E1C4}0.348          & \cellcolor[HTML]{A0D7AE}0.471          & \cellcolor[HTML]{DDF1E2}0.169          & \cellcolor[HTML]{B5E0C1}0.366          & \cellcolor[HTML]{A9DAB6}0.427          & \cellcolor[HTML]{9AD4A9}0.500          & \cellcolor[HTML]{83CA96}0.613          & \cellcolor[HTML]{84CB97}\textbf{0.607} & \cellcolor[HTML]{68BF7F}\textbf{0.745} & \cellcolor[HTML]{C1E5CB}0.306          & \cellcolor[HTML]{B3DFBF}0.375          \\
\textbf{ROUGE-L-Re\textsuperscript{*}}       & \cellcolor[HTML]{ACDCB9}0.409          & \cellcolor[HTML]{83CA96}0.614          & \cellcolor[HTML]{C3E5CC}\textbf{0.300} & \cellcolor[HTML]{96D2A6}0.520          & \cellcolor[HTML]{90D0A1}0.551          & \cellcolor[HTML]{7EC891}0.640          & \cellcolor[HTML]{75C48B}0.680          & \cellcolor[HTML]{B4DFBF}0.374          & \cellcolor[HTML]{ABDBB8}0.416          & \cellcolor[HTML]{7AC68E}0.660          & \cellcolor[HTML]{74C489}\textbf{0.688} \\
\textbf{ROUGE-L-F1\textsuperscript{*}}       & \cellcolor[HTML]{B2DEBD}0.384          & \cellcolor[HTML]{7CC790}0.646          & \cellcolor[HTML]{C6E7CE}0.285          & \cellcolor[HTML]{92D1A3}0.538          & \cellcolor[HTML]{8DCE9E}0.564          & \cellcolor[HTML]{79C68E}0.661          & \cellcolor[HTML]{6EC184}0.719          & \cellcolor[HTML]{9ED6AD}0.479          & \cellcolor[HTML]{93D1A4}0.534          & \cellcolor[HTML]{84CA97}0.610          & \cellcolor[HTML]{7BC78F}0.655          \\
\textbf{CHRF\textsuperscript{*}}             & \cellcolor[HTML]{BAE2C5}0.341          & \cellcolor[HTML]{A2D7B0}0.460          & \cellcolor[HTML]{FFFFFF}-0.075         & \cellcolor[HTML]{A2D7B0}0.463          & \cellcolor[HTML]{A7D9B4}0.438          & \cellcolor[HTML]{8ACD9C}0.581          & \cellcolor[HTML]{8ECF9F}0.560          & \cellcolor[HTML]{99D4A9}0.504          & \cellcolor[HTML]{9DD5AC}0.484          & \cellcolor[HTML]{9DD5AC}0.483          & \cellcolor[HTML]{A2D7B0}0.462          \\
\textbf{METEOR\textsuperscript{*}}           & \cellcolor[HTML]{ABDBB8}\textbf{0.415} & \cellcolor[HTML]{78C58D}\textbf{0.667} & \cellcolor[HTML]{D6EEDD}0.203          & \cellcolor[HTML]{94D1A5}0.529          & \cellcolor[HTML]{8ACD9C}\textbf{0.581} & \cellcolor[HTML]{77C58C}0.674          & \cellcolor[HTML]{6FC185}0.713          & \cellcolor[HTML]{A8DAB6}0.429          & \cellcolor[HTML]{A2D7B0}0.463          & \cellcolor[HTML]{78C58D}\textbf{0.668} & \cellcolor[HTML]{72C387}\textbf{0.699} \\
\textbf{BLEU\textsuperscript{*}}             & \cellcolor[HTML]{B2DEBE}0.382          & \cellcolor[HTML]{7DC891}0.642          & \cellcolor[HTML]{ECF7EF}0.098          & \cellcolor[HTML]{8ECFA0}0.557          & \cellcolor[HTML]{8DCE9E}0.565          & \cellcolor[HTML]{72C387}0.698          & \cellcolor[HTML]{71C287}0.702          & \cellcolor[HTML]{A5D9B3}0.447          & \cellcolor[HTML]{A4D8B2}0.453          & \cellcolor[HTML]{74C48A}\textbf{0.685} & \cellcolor[HTML]{74C48A}0.686          \\\hline
\textbf{Levenshtein dist.} & \cellcolor[HTML]{90D0A1}\textbf{0.547} & \cellcolor[HTML]{61BC79}\textbf{0.780} & \cellcolor[HTML]{A4D8B2}\textbf{0.453} & \cellcolor[HTML]{86CB98}\textbf{0.600} & \cellcolor[HTML]{7BC78F}\textbf{0.654} & \cellcolor[HTML]{7CC790}0.650          & \cellcolor[HTML]{65BD7D}\textbf{0.760} & \cellcolor[HTML]{8DCE9E}0.566          & \cellcolor[HTML]{8FCFA0}0.555          & \cellcolor[HTML]{94D1A4}0.531          & \cellcolor[HTML]{72C388}\textbf{0.697} \\
\textbf{WER}               & \cellcolor[HTML]{CFEBD6}0.239          & \cellcolor[HTML]{80C993}0.629          & \cellcolor[HTML]{F4FAF5}0.059          & \cellcolor[HTML]{BDE3C7}0.326          & \cellcolor[HTML]{90D0A1}0.550          & \cellcolor[HTML]{ACDCB9}0.412          & \cellcolor[HTML]{71C286}0.704          & \cellcolor[HTML]{9AD4AA}0.499          & \cellcolor[HTML]{93D1A3}0.535          & \cellcolor[HTML]{CCEAD4}0.252          & \cellcolor[HTML]{7FC993}0.631          \\
\textbf{MER}               & \cellcolor[HTML]{B0DDBC}0.392          & \cellcolor[HTML]{7FC892}0.635          & \cellcolor[HTML]{E0F2E5}0.156          & \cellcolor[HTML]{8DCE9E}0.565          & \cellcolor[HTML]{8ECFA0}0.557          & \cellcolor[HTML]{71C287}0.703          & \cellcolor[HTML]{70C286}0.706          & \cellcolor[HTML]{9AD4A9}0.500          & \cellcolor[HTML]{97D3A7}0.513          & \cellcolor[HTML]{7AC68E}0.659          & \cellcolor[HTML]{7BC790}0.651          \\
\textbf{WIL}               & \cellcolor[HTML]{B0DDBC}0.394          & \cellcolor[HTML]{7CC790}0.649          & \cellcolor[HTML]{E8F5EB}0.117          & \cellcolor[HTML]{88CC9A}\textbf{0.590} & \cellcolor[HTML]{8DCE9E}0.566          & \cellcolor[HTML]{68BF7F}\textbf{0.747} & \cellcolor[HTML]{6DC183}0.723          & \cellcolor[HTML]{8ACD9C}\textbf{0.578} & \cellcolor[HTML]{8DCE9E}0.566          & \cellcolor[HTML]{78C58D}\textbf{0.668} & \cellcolor[HTML]{7EC892}0.638          \\\hline
\textbf{ROUGE-WE\textsuperscript{*}}         & \cellcolor[HTML]{AEDDBA}0.402          & \cellcolor[HTML]{81C994}0.624          & \cellcolor[HTML]{DEF1E3}0.165          & \cellcolor[HTML]{9BD4AA}0.496          & \cellcolor[HTML]{90D0A1}0.549          & \cellcolor[HTML]{81C995}0.621          & \cellcolor[HTML]{6FC285}0.712          & \cellcolor[HTML]{ABDBB8}0.415          & \cellcolor[HTML]{95D2A5}0.524          & \cellcolor[HTML]{87CC99}0.595          & \cellcolor[HTML]{7CC790}0.650          \\
\textbf{SkipThoughts\textsuperscript{*}}     & \cellcolor[HTML]{C3E6CC}0.298          & \cellcolor[HTML]{AEDCBA}0.403          & \cellcolor[HTML]{FFFFFF}-0.067         & \cellcolor[HTML]{D1ECD8}0.229          & \cellcolor[HTML]{B3DFBF}0.375          & \cellcolor[HTML]{B5E0C1}0.366          & \cellcolor[HTML]{99D4A9}0.504          & \cellcolor[HTML]{BBE2C5}0.338          & \cellcolor[HTML]{ADDCB9}0.407          & \cellcolor[HTML]{C5E6CE}0.288          & \cellcolor[HTML]{A6D9B4}0.439          \\
\textbf{Embedding Avg\textsuperscript{*}}    & \cellcolor[HTML]{CAE8D2}0.266          & \cellcolor[HTML]{B3DFBF}0.375          & \cellcolor[HTML]{FFFFFF}-0.209         & \cellcolor[HTML]{F3FAF4}0.064          & \cellcolor[HTML]{ACDCB9}0.412          & \cellcolor[HTML]{D2ECD9}0.223          & \cellcolor[HTML]{8BCE9D}0.572          & \cellcolor[HTML]{E2F3E6}0.147          & \cellcolor[HTML]{B0DDBC}0.392          & \cellcolor[HTML]{D5EDDB}0.211          & \cellcolor[HTML]{91D0A2}0.543          \\
\textbf{VectorExtrema\textsuperscript{*}}    & \cellcolor[HTML]{ACDCB9}0.409          & \cellcolor[HTML]{8FCFA0}0.553          & \cellcolor[HTML]{E6F4EA}0.127          & \cellcolor[HTML]{A9DBB7}0.424          & \cellcolor[HTML]{9AD4A9}0.500          & \cellcolor[HTML]{90D0A1}0.550          & \cellcolor[HTML]{7CC790}0.648          & \cellcolor[HTML]{B5E0C0}0.367          & \cellcolor[HTML]{A0D7AF}0.468          & \cellcolor[HTML]{94D1A4}0.531          & \cellcolor[HTML]{86CB98}0.600          \\
\textbf{GreedyMatching\textsuperscript{*}}   & \cellcolor[HTML]{C1E5CB}0.308          & \cellcolor[HTML]{8ACD9C}0.577          & \cellcolor[HTML]{FFFFFF}\red{-0.041}         & \cellcolor[HTML]{C4E6CD}0.295          & \cellcolor[HTML]{96D2A6}0.520          & \cellcolor[HTML]{A7DAB5}0.436          & \cellcolor[HTML]{77C58C}0.670          & \cellcolor[HTML]{C6E7CF}0.281          & \cellcolor[HTML]{9ED6AD}0.479          & \cellcolor[HTML]{A9DAB6}0.428          & \cellcolor[HTML]{81C994}0.624          \\
\textbf{USE\textsuperscript{*}}              & \cellcolor[HTML]{BBE2C5}0.339          & \cellcolor[HTML]{96D2A6}0.522          & \cellcolor[HTML]{D7EEDD}0.201          & \cellcolor[HTML]{B5E0C1}0.366          & \cellcolor[HTML]{9FD6AE}0.476          & \cellcolor[HTML]{9FD6AE}0.474          & \cellcolor[HTML]{7EC892}0.637          & \cellcolor[HTML]{BDE3C7}0.327          & \cellcolor[HTML]{A4D8B2}0.452          & \cellcolor[HTML]{A3D8B1}0.454          & \cellcolor[HTML]{86CB99}0.598          \\
\textbf{WMD}               & \cellcolor[HTML]{B8E1C3}0.354          & \cellcolor[HTML]{87CC99}0.594          & \cellcolor[HTML]{E0F2E5}0.154          & \cellcolor[HTML]{AADBB7}0.421          & \cellcolor[HTML]{94D1A5}0.529          & \cellcolor[HTML]{8ECF9F}0.561          & \cellcolor[HTML]{77C58C}0.670          & \cellcolor[HTML]{BFE4C9}0.319          & \cellcolor[HTML]{ABDBB8}0.414          & \cellcolor[HTML]{8ACD9C}0.577          & \cellcolor[HTML]{77C58C}0.670          \\
\textbf{BertScore\textsuperscript{*}}        & \cellcolor[HTML]{9BD4AA}\textbf{0.497} & \cellcolor[HTML]{74C489}\textbf{0.688} & \cellcolor[HTML]{BAE2C5}\textbf{0.340} & \cellcolor[HTML]{8CCE9D}0.571          & \cellcolor[HTML]{88CC9A}\textbf{0.590} & \cellcolor[HTML]{6FC285}0.710          & \cellcolor[HTML]{68BF80}\textbf{0.744} & \cellcolor[HTML]{94D1A4}0.530          & \cellcolor[HTML]{8FCFA1}0.552          & \cellcolor[HTML]{7DC791}0.645          & \cellcolor[HTML]{76C58B}0.676          \\
\textbf{MoverScore\textsuperscript{*}}       & \cellcolor[HTML]{B6E0C2}0.360          & \cellcolor[HTML]{7EC891}0.640          & \cellcolor[HTML]{CEEAD5}0.246          & \cellcolor[HTML]{8CCE9D}0.570          & \cellcolor[HTML]{8ECF9F}0.559          & \cellcolor[HTML]{74C489}0.687          & \cellcolor[HTML]{74C489}0.688          & \cellcolor[HTML]{A5D9B2}0.448          & \cellcolor[HTML]{A1D7AF}0.467          & \cellcolor[HTML]{78C58C}\textbf{0.669} & \cellcolor[HTML]{7AC68E}0.657          \\\hline
\textbf{Stanza+Snomed\textsuperscript{*}}    & \cellcolor[HTML]{BCE2C6}0.334          & \cellcolor[HTML]{98D3A8}0.508          & \cellcolor[HTML]{E8F5EB}0.118          & \cellcolor[HTML]{B8E1C3}0.354          & \cellcolor[HTML]{A2D7B0}0.460          & \cellcolor[HTML]{94D2A5}0.528          & \cellcolor[HTML]{7DC891}0.643          & \cellcolor[HTML]{A5D9B3}0.447          & \cellcolor[HTML]{93D1A4}0.533          & \cellcolor[HTML]{A4D8B2}0.449          & \cellcolor[HTML]{8FCFA0}0.553\\

    \end{tabular}
    \caption{Spearman's correlation coefficients for each metric and each criterion. In bold are the top three scores per column. \wip{All the metrics marked with an asterisk (\textsuperscript{*}) are inversely correlated with the given criterion (e.g. higher post-edit time means worse note, but higher ROUGE score means better note)}; the sign of the coefficient is inverted for ease of visualisation. Coefficients in red are not statistically significant (with p > 0.05).}
    \label{tab:metrics-correlation}
\end{table*}

Borrowing from the field of Automatic Summarisation, most studies on Note Generation rely on ROUGE and fact-extraction based metrics to evaluate the generated notes (Section \ref{sec:related-work} for more details). While some studies carry out a small human evaluation, there is little effort to investigate whether the scores from ROUGE or the other metrics employed correlate well with the human judgements, especially extrinsic criteria such as post-edit times. However, scores from these  metrics are featured on leaderboards\footnote{\url{https://nlpprogress.com/english/summarization.html}} for summarisation tasks, driving future research.
To address this, we carry out a correlation study of automatic metrics for the task of Note Generation. A total of 18 automatic metrics are tested against statistics produced by the human judgements of our criteria: post-edit times, number of incorrect statements, and number of omissions. Following the taxonomies reported by \citet{celikyilmaz2020evaluation} and \citet{sai2020survey}, the metrics considered can be loosely grouped in:
\begin{itemize}
    \item \textbf{Text overlap metrics.} These are based on string matching, whether character based, word based, or n-gram based. Some use stemming, synonyms, or paraphrases. They include: ROUGE \cite{lin2004rouge}, CHRF \cite{popovic2015chrf}, METEOR \cite{lavie2007meteor}, and BLEU \cite{papineni2002bleu}.
    
    \item \textbf{Edit distance metrics.} These count the number of character or word level transformations required to convert the system output into the reference text. They include: Levenshtein edit distance \cite{levenshtein1966binary}, WER \cite{su1992new}, MER and WIL \cite{Morris2004FromWA}. 
    
    \item \textbf{Embedding metrics,} including word-level, byte-level, and sentence-level embeddings. These metrics encode units of text with pretrained models and compute cosine similarity between them. They include: ROUGE-WE \cite{Morris2004FromWA}, SkipThoughts, EmbeddingAverage, VectorExtrema \cite{forgues2014bootstrapping}, GreedyMatching \cite{sharma2017nlgeval}, USE\footnote{Cosine similarity between the reference and the hypothesis embeddings. Embeddings computed with Universal Sentence Encoder.} \cite{cer2018universal}, WMD \cite{kusner2015word}, BertScore \cite{zhang2019bertscore}, and MoverScore \cite{zhao2019moverscore}.
    
    \item \textbf{Fact-extraction.} The Stanza+Snomed metric extracts medical concept spans with Stanza \citep{zhang2021biomedical}, then uses similarity measures to map them to entities in the SNOMED CT clinical ontology \cite{spackman1997snomed}. The metric computes F1 score between reference and hypothesis over the set of extracted entities. 
\end{itemize}

For more details on each metric please refer to their respective papers. All these metrics attempt to measure the accuracy of the generated text by comparing it against a reference text. Our human evaluation study produces three distinct human-curated notes which can be used as reference: the \emph{human\_note} is the original note, written by the consulting clinician (and also one of the hypotheses), the \emph{eval\_note} is the note written by the evaluators after listening to the consultation audio, and the \emph{edited\_note} is the generated note after being post-edited by the evaluators. \wip{Table \ref{tab:metrics-correlation} reports the correlation coefficients. When correlating against post-edit times, we consider each reference text (\emph{human\_note}, \emph{edited\_note}, \emph{eval\_note}) separately, then take the average and the maximum of the metric scores for each reference. For count of incorrect statements, omissions, and incorrect+omissions we only report the average and the maximum scores, taking all three references into account as commonly done by the metrics that support multiple references (e.g. BLEU, ROUGE, METEOR, BertScore).
We compute Pearson's and Spearman's coefficients and, upon finding similar patterns, only report Spearman's coefficients in Table \ref{tab:metrics-correlation}. The Pearson's coefficients can be found in Table \ref{tab:metrics-correlation-pearson} in the Appendix.}

\wip{As shown in Table \ref{tab:metrics-correlation}, all metrics display a strong bias} towards the choice of reference. In particular, the correlation scores with the \emph{edited\_note} as reference are much higher than those of either \emph{human\_note} or \emph{eval\_note}. As the \emph{edited\_note} is a transformation of the generated note (refer to Figure \ref{fig:edit-note}), these high correlations show how reliant all the metrics are on the surface form of the text. 
The significant difference between taking \emph{human\_note} and \emph{eval\_note} as reference can be traced to two main factors: (i) the \emph{human\_note} is unique per consultation so the human judgements are averaged across evaluators (reducing noise and collapsing disagreement), and (ii) the \emph{eval\_note} was not written to replace a SOAP note but is rather a list of the most salient points in the consultation, and sometimes contains more information than would typically be detailed in a SOAP note.

The top three metrics in most scenarios are Levenshtein distance, BertScore, and METEOR. While METEOR and BertScore are established metrics in NLG evaluation, Levenshtein distance is not typically used as a metric in long-form text evaluation. From a semantic point of view, edit distance has the least amount of knowledge and should be very brittle when comparing text that is meaningfully similar but lexically very different. Yet Levenshtein distance has the highest correlation even when the reference is the \emph{eval\_note}, which is syntactically very different from the generated note; whereas even contextual metrics like BertScore perform more poorly. 
A possible explanation for this behaviour may be that our post-editing times and count of incorrect statements/omissions---unlike Likert scales scores---measure the amount of work required to convert a synthetic note into a factually-correct and relevant note, just as Levenshtein distance measures the character-level distance between the synthetic note and the reference.

We notice that all the metrics correlate better with counts of \textit{incorrect+omissions} than with post-edit times, despite the two criteria being strongly correlated with each other (0.829 Spearman's correlation, see Table \ref{tab:criteria-correlation}). We believe this is due to post-editing times containing more noise and capturing more of the stylistic differences between evaluators than the number of errors does. 

\wip{Correlation matrices between the metrics scores can be found in Appendix \ref{app:correlation-matrices}.}
\section{Conclusions}



\wip{We conducted a human evaluation study for the task of consultation Note Generation, computed agreement between evaluators, and quantified the extent to which human error impacts the judgements. We then carried out a correlation study with 18 automatic metrics, discussing their limitations and identifying the most successful ones.}

We found that the choice of human reference has a significant effect on all automatic metrics and that simple character-based metrics like Levenshtein distance can be more effective than complex model-based metrics for the task of Note Generation. \wip{Based on our findings, character-based Levenshtein distance, BertScore, and METEOR are the most suitable metrics to evaluate this task.} We release all the data and annotations and welcome researchers to assess further metrics.

\section{Acknowledgements}
The authors would like to thank Rachel Young and Tom Knoll for supporting the team and hiring the evaluators, Vitalii Zhelezniak for his advice on revising the paper, and Kristian Boda for helping to set up the Stanza+Snomed fact-extraction system.

\bibliography{anthology,custom}
\bibliographystyle{acl_natbib}

\newpage

\appendix
\renewcommand\thefigure{A.\arabic{figure}} 
\renewcommand\thetable{A.\arabic{table}} 
\section*{Appendix}




\begin{figure*}[t]
    \centering
    \includegraphics[width=1.0\textwidth]{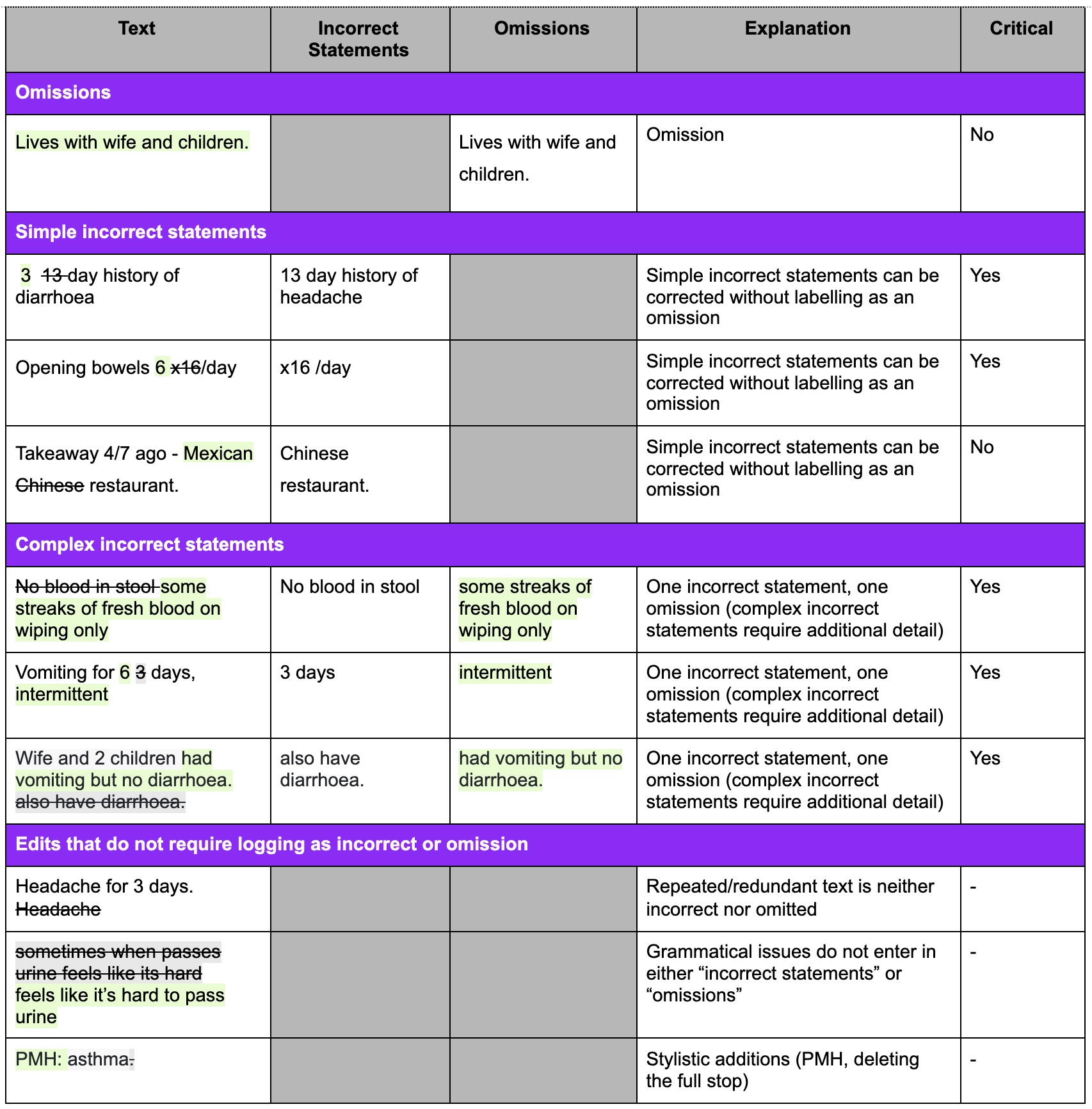}
    \caption{Table of examples given to the evaluators for reference.}
    \label{fig:evaluators-instructions}
\end{figure*}

\section{Instructions for evaluators}
\label{app:instructions}
Figure \ref{fig:evaluators-instructions} shows a table of examples provided to the evaluators to help them understand how to list incorrect statements and omissions, especially around edge cases.


\section{Pairwise Agreement}
\label{app:pairwise-agreement}
Pairwise agreement is reported in Table \ref{tab:pairwise-agreement}. The agreement on post-edit times is a rank agreement computed by ranking the times and using ordinal Krippendorff Alpha; the number of incorrect statements and number of omissions agreement is computed with interval Krippendorff Alpha; and the word overlap is computed with word-level F1 score.

\begin{table*}[t]
    \centering
    \begin{tabular}{p{1.1cm}|p{1.1cm}|p{2.0cm}|p{1.5cm}|p{2.4cm}|p{1.5cm}|p{2.4cm}}
        \textbf{Eval 1} & \textbf{Eval 2} & \textbf{Post-edit times} & \multicolumn{2}{c|}{\textbf{Incorrect statements}} & \multicolumn{2}{c}{\textbf{Omissions}} \\\hline
        & & & \textbf{Count} & \textbf{Word Overlap} & \textbf{Count} & \textbf{Word Overlap}\\\hline
        eval1 & eval2 & 0.444 & 0.573 & 0.369 & 0.124 & 0.240\\
        eval1 & eval3 & 0.534 & 0.599 & 0.484 & 0.452 & 0.266\\
        eval1 & eval4 & 0.660 & 0.624 & 0.471 & 0.538 & 0.282\\
        eval1 & eval5 & 0.591 & 0.639 & 0.415 & 0.408 & 0.260\\
        eval2 & eval3 & 0.408 & 0.413 & 0.386 & 0.232 & 0.240\\
        eval2 & eval4 & 0.420 & 0.303 & 0.389 & -0.024 & 0.243\\
        eval2 & eval5 & 0.634 & 0.717 & 0.401 & 0.543 & 0.270\\
        eval3 & eval4 & 0.501 & 0.626 & 0.531 & 0.449 & 0.307\\
        eval3 & eval5 & 0.520 & 0.512 & 0.433 & 0.495 & 0.269 \\
        eval4 & eval5 & 0.664 & 0.371 & 0.436 & 0.294 & 0.301 \\
    \end{tabular}
    \caption{Pairwise agreement between evaluators.}
    \label{tab:pairwise-agreement}
\end{table*}


\section{Human error examples}
\label{app:human-errors}

\large{Example human note 1.}
\begin{Verbatim}[commandchars=\\\{\},fontsize=\small]
3/7 hx of diarrhea, mainly watery. 
No blood in stool. 
Opening bowels x6/day. 
Associated LLQ pain - crampy, inter-
mittent, nil radiation. 
Also vomiting - \textbf{\textcolor{red}{mainly bilous}}. 
No blood in vomit. 
Fever on first day, nil since. 
Has been feeling lethargic and weak since. 
Takeaway 4/7 ago - Chinese restaurant. 
\textbf{\textcolor{red}{Wife and children also unwell}} 
\textbf{\textcolor{red}{with vomiting, but no diarrhea.}} 
No other unwell contacts. 
PMH: Asthma
DH: Inhalers
SH: works as an accountant. 
Lives with wife and children. 
Affecting his ADLs as has to be near 
toilet often. 
Nil smoking/etOH hx
\end{Verbatim}

Incorrect statements
\begin{itemize}
    \item Also vomiting - \sout{mainly bilous}, resolved after 1st day \textcolor{blue}{(complex incorrect statement, critical)}
    \item \sout{Wife and children also unwell with vomiting}. -> 1 child had some vomiting but no other symptoms in wife and other child. \textcolor{blue}{(complex incorrect statement, non-critical)}
\end{itemize}

\large{Example human note 2.}
\begin{Verbatim}[fontsize=\small]
PC: Cough and cold.
HPC: 4-5 day hx runny nose, dry cough.
No sputum/haemoptysis. 
No epistaxis/sinus pain. 
Feels hot, hasn't measured temperature. 
No SOB/inspiratory pain/wheeze. 
Aches and pains. No vomiting. 
E&D ok. PUing ok. 
No hx chest problems/recurrent chest
infections/wt loss.
Thinks last BP was fine, can't remember
when last BP/DM check up was.
Doesn't check blood sugars/urine at home. 
No increased thirst/urinary freq.
Taking 2-3 lemsips a day which eases sx.
PMH: Hypertension. T2 DM.
DH: Lisinopril. Metformin
Height- 5ft 5in
Weight - 65kg
SH: Non smoker, odd sherry. 
Lives with partner and dog. 
Office manager.
\end{Verbatim}

Here the Lead Clinician found no incorrect statements. However, some of the evaluators did.

\vspace{0.2cm}

\begin{tabular}{p{.5cm}|p{5.7cm}}
     E1 & -- dm \\\hline
     E2 & \\\hline
     \multirow{2}{*}{E3} & -- Thinks last BP was fine, can't remember when last BP/DM check up was.\\
      &  -- Doesn't check blood sugars/urine at home. \\\hline
     \multirow{2}{*}{E4} & -- can't remember when last DM check up was.\\
        & -- Doesn't check blood sugars/urine at home.\\
        & -- 4 day hx\\\hline
     E5 & \\
\end{tabular}

\vspace{0.2cm}

Upon inspecting the recording of the consultation, the Lead Clinician found that the words `Thinks last BP was fine, can’t remember when last BP/DM check up was. Doesn’t check blood sugars/urine at home' were not stated by the patient. Instead, the patient made a tangential statement summarised here: `diabetes (generally well controlled, last blood test was 3 weeks ago), hypertension (doesn’t remember last blood pressure check)'. 
Please note that in diabetic and hypertension patients, clinical convention is to indicate severity of diagnosis by whether the patient requires home monitoring. Therefore, the clinician inferred part of the statement. Furthermore, clinical convention is to differentiate diabetes mellitus (DM) from diabetes insipidus (DI). Therefore again, the clinician inferred part of the statement. Any other errors are attributed to normal human error due excessive patient detail (non-critical).

\begin{table*}[t]
    \setlength{\tabcolsep}{3.2pt} 
    \centering
    \def\arraystretch{.9} 
    \begin{tabular}{l|ccc|c|c|c|c|c|c|c|c}
         \textbf{Human scores:} & \multicolumn{5}{c|}{\textbf{Timings}} & \multicolumn{2}{c|}{\textbf{Inc+Omi}} & \multicolumn{2}{c|}{\textbf{Inc}} & \multicolumn{2}{c}{\textbf{Omi}}\\\hline
         \textbf{Reference:} & \textbf{human}  & \textbf{edited} & \textbf{eval} & \textbf{avg} & \textbf{max} & \textbf{avg} & \textbf{max} & \textbf{avg} & \textbf{max} & \textbf{avg} & \textbf{max}\\\hline

\textbf{ROUGE-1-F1*}       & \cellcolor[HTML]{B4DFBF}0.374          & \cellcolor[HTML]{89CD9B}0.583          & \cellcolor[HTML]{E0F2E5}0.156          & \cellcolor[HTML]{ABDBB8}0.417          & \cellcolor[HTML]{92D1A3}0.539          & \cellcolor[HTML]{8BCE9D}0.574          & \cellcolor[HTML]{73C388}0.693          & \cellcolor[HTML]{ACDCB8}0.413          & \cellcolor[HTML]{9FD6AE}0.474          & \cellcolor[HTML]{92D1A3}0.537          & \cellcolor[HTML]{78C58D}0.667          \\
\textbf{ROUGE-2-F1*}       & \cellcolor[HTML]{B3DFBE}0.378          & \cellcolor[HTML]{85CB98}0.601          & \cellcolor[HTML]{E1F3E6}0.149          & \cellcolor[HTML]{A9DAB6}0.426          & \cellcolor[HTML]{91D0A2}0.545          & \cellcolor[HTML]{86CC99}0.597          & \cellcolor[HTML]{6EC184}\textbf{0.717} & \cellcolor[HTML]{9DD5AC}0.484          & \cellcolor[HTML]{92D1A3}0.538          & \cellcolor[HTML]{96D2A6}0.519          & \cellcolor[HTML]{7BC78F}0.655          \\
\textbf{ROUGE-3-F1*}       & \cellcolor[HTML]{B5DFC0}0.368          & \cellcolor[HTML]{87CC99}0.594          & \cellcolor[HTML]{EAF6ED}0.105          & \cellcolor[HTML]{ABDBB8}0.417          & \cellcolor[HTML]{93D1A4}0.533          & \cellcolor[HTML]{86CB98}0.600          & \cellcolor[HTML]{6FC285}0.711          & \cellcolor[HTML]{9AD4A9}0.501          & \cellcolor[HTML]{90D0A1}0.550          & \cellcolor[HTML]{98D3A8}0.511          & \cellcolor[HTML]{7EC892}0.638          \\
\textbf{ROUGE-4-F1*}       & \cellcolor[HTML]{B6E0C1}0.363          & \cellcolor[HTML]{89CD9B}0.583          & \cellcolor[HTML]{EFF8F1}0.082          & \cellcolor[HTML]{ACDCB8}0.413          & \cellcolor[HTML]{96D2A6}0.518          & \cellcolor[HTML]{85CB98}0.602          & \cellcolor[HTML]{71C287}0.702          & \cellcolor[HTML]{98D3A8}0.510          & \cellcolor[HTML]{8FCFA0}\textbf{0.553} & \cellcolor[HTML]{99D3A8}0.507          & \cellcolor[HTML]{81C995}0.621          \\
\textbf{ROUGE-L-Pr*}       & \cellcolor[HTML]{B5DFC0}0.368          & \cellcolor[HTML]{B4DFC0}0.370          & \cellcolor[HTML]{E6F5EA}0.125          & \cellcolor[HTML]{C0E4C9}0.315          & \cellcolor[HTML]{B5DFC0}0.368          & \cellcolor[HTML]{9CD5AB}0.492          & \cellcolor[HTML]{88CC9A}0.589          & \cellcolor[HTML]{90D0A1}\textbf{0.547} & \cellcolor[HTML]{74C48A}\textbf{0.685} & \cellcolor[HTML]{BFE4C9}0.318          & \cellcolor[HTML]{B6E0C2}0.36           \\
\textbf{ROUGE-L-Re*}       & \cellcolor[HTML]{ADDCBA}0.406          & \cellcolor[HTML]{87CC99}0.594          & \cellcolor[HTML]{C1E5CB}\textbf{0.308} & \cellcolor[HTML]{9CD5AB}\textbf{0.491} & \cellcolor[HTML]{8FCFA0}\textbf{0.554} & \cellcolor[HTML]{80C994}\textbf{0.627} & \cellcolor[HTML]{77C58C}0.674          & \cellcolor[HTML]{B1DEBD}0.389          & \cellcolor[HTML]{B2DEBE}0.381          & \cellcolor[HTML]{7FC993}\textbf{0.632} & \cellcolor[HTML]{70C286}\textbf{0.708} \\
\textbf{ROUGE-L-F1*}       & \cellcolor[HTML]{B1DEBD}0.387          & \cellcolor[HTML]{87CC99}0.595          & \cellcolor[HTML]{C6E7CF}0.283          & \cellcolor[HTML]{A5D9B3}0.445          & \cellcolor[HTML]{8FCFA0}0.553          & \cellcolor[HTML]{85CB98}0.603          & \cellcolor[HTML]{6FC185}0.714          & \cellcolor[HTML]{9DD5AC}0.483          & \cellcolor[HTML]{97D3A7}0.513          & \cellcolor[HTML]{94D1A5}0.529          & \cellcolor[HTML]{78C58D}0.668          \\
\textbf{CHRF*}             & \cellcolor[HTML]{B5DFC0}0.368          & \cellcolor[HTML]{A5D9B3}0.445          & \cellcolor[HTML]{FFFFFF}\red{-0.013}         & \cellcolor[HTML]{AFDDBB}0.397          & \cellcolor[HTML]{ADDCBA}0.404          & \cellcolor[HTML]{90D0A1}0.547          & \cellcolor[HTML]{91D0A2}0.542          & \cellcolor[HTML]{9FD6AE}0.476          & \cellcolor[HTML]{A0D7AF}0.469          & \cellcolor[HTML]{A4D8B2}0.451          & \cellcolor[HTML]{A4D8B2}0.45           \\
\textbf{METEOR*}           & \cellcolor[HTML]{B3DFBE}0.379          & \cellcolor[HTML]{82CA95}\textbf{0.617} & \cellcolor[HTML]{D5EDDB}0.210          & \cellcolor[HTML]{ABDBB8}0.414          & \cellcolor[HTML]{A8DAB5}0.433          & \cellcolor[HTML]{8ACD9C}0.580          & \cellcolor[HTML]{88CC9A}0.591          & \cellcolor[HTML]{A8DAB5}0.431          & \cellcolor[HTML]{A5D9B3}0.447          & \cellcolor[HTML]{93D1A4}0.533          & \cellcolor[HTML]{92D1A3}0.537          \\
\textbf{BLEU*}             & \cellcolor[HTML]{B4DFC0}0.370          & \cellcolor[HTML]{84CB97}0.609          & \cellcolor[HTML]{E8F5EC}0.115          & \cellcolor[HTML]{A6D9B3}0.443          & \cellcolor[HTML]{90D0A1}0.551          & \cellcolor[HTML]{85CB98}0.603          & \cellcolor[HTML]{76C58B}0.678          & \cellcolor[HTML]{A6D9B4}0.440          & \cellcolor[HTML]{ACDCB9}0.409          & \cellcolor[HTML]{8ECF9F}0.560          & \cellcolor[HTML]{73C388}\textbf{0.693} \\
\textbf{Levenshtein dist.} & \cellcolor[HTML]{99D4A9}\textbf{0.503} & \cellcolor[HTML]{63BC7B}\textbf{0.772} & \cellcolor[HTML]{A0D6AE}\textbf{0.472} & \cellcolor[HTML]{83CA96}\textbf{0.612} & \cellcolor[HTML]{78C58D}\textbf{0.667} & \cellcolor[HTML]{79C68D}\textbf{0.663} & \cellcolor[HTML]{64BD7C}\textbf{0.765} & \cellcolor[HTML]{90D0A1}\textbf{0.550} & \cellcolor[HTML]{90D0A1}0.547          & \cellcolor[HTML]{8CCE9E}\textbf{0.567} & \cellcolor[HTML]{6EC184}\textbf{0.718} \\
\textbf{WER}               & \cellcolor[HTML]{C7E7CF}0.280          & \cellcolor[HTML]{8DCE9E}0.565          & \cellcolor[HTML]{F8FCF9}\red{0.035}          & \cellcolor[HTML]{C4E6CD}0.291          & \cellcolor[HTML]{97D3A7}0.515          & \cellcolor[HTML]{ABDBB8}0.416          & \cellcolor[HTML]{74C489}0.687          & \cellcolor[HTML]{9FD6AD}0.477          & \cellcolor[HTML]{95D2A5}0.527          & \cellcolor[HTML]{CBE9D3}0.258          & \cellcolor[HTML]{82CA95}0.62           \\
\textbf{MER}               & \cellcolor[HTML]{B3DFBF}0.377          & \cellcolor[HTML]{88CC9A}0.590          & \cellcolor[HTML]{E2F3E7}0.144          & \cellcolor[HTML]{A9DAB6}0.427          & \cellcolor[HTML]{94D1A4}0.531          & \cellcolor[HTML]{86CB98}0.600          & \cellcolor[HTML]{73C388}0.693          & \cellcolor[HTML]{9DD5AC}0.485          & \cellcolor[HTML]{9BD4AA}0.497          & \cellcolor[HTML]{96D2A6}0.522          & \cellcolor[HTML]{7CC790}0.649          \\
\textbf{WIL}               & \cellcolor[HTML]{B6E0C1}0.363          & \cellcolor[HTML]{86CB98}0.599          & \cellcolor[HTML]{EFF9F2}0.080          & \cellcolor[HTML]{A8DAB6}0.429          & \cellcolor[HTML]{93D1A4}0.534          & \cellcolor[HTML]{83CA96}0.615          & \cellcolor[HTML]{70C286}0.707          & \cellcolor[HTML]{96D2A6}\textbf{0.522} & \cellcolor[HTML]{8ECFA0}\textbf{0.557} & \cellcolor[HTML]{97D3A7}0.517          & \cellcolor[HTML]{80C994}0.626          \\
\textbf{ROUGE-WE*}         & \cellcolor[HTML]{B0DDBC}0.392          & \cellcolor[HTML]{8BCE9D}0.573          & \cellcolor[HTML]{E3F3E8}0.139          & \cellcolor[HTML]{A8DAB5}0.432          & \cellcolor[HTML]{94D1A4}0.530          & \cellcolor[HTML]{89CD9B}0.586          & \cellcolor[HTML]{72C387}0.698          & \cellcolor[HTML]{A6D9B4}0.440          & \cellcolor[HTML]{9AD4AA}0.498          & \cellcolor[HTML]{93D1A3}0.535          & \cellcolor[HTML]{7AC68F}0.656          \\
\textbf{SkipThoughts*}     & \cellcolor[HTML]{BEE3C8}0.323          & \cellcolor[HTML]{C2E5CC}0.301          & \cellcolor[HTML]{FFFFFF}-0.099         & \cellcolor[HTML]{D4EDDB}0.213          & \cellcolor[HTML]{C2E5CC}0.301          & \cellcolor[HTML]{BBE2C5}0.338          & \cellcolor[HTML]{AEDCBA}0.403          & \cellcolor[HTML]{C2E5CB}0.305          & \cellcolor[HTML]{C3E6CC}0.298          & \cellcolor[HTML]{C9E8D1}0.270          & \cellcolor[HTML]{B4DFC0}0.371          \\
\textbf{Embedding Avg*}    & \cellcolor[HTML]{D1EBD8}0.230          & \cellcolor[HTML]{CBE9D3}0.260          & \cellcolor[HTML]{FFFFFF}-0.188         & \cellcolor[HTML]{F8FCF9}\red{0.039}          & \cellcolor[HTML]{C2E5CB}0.304          & \cellcolor[HTML]{DBF0E1}0.180          & \cellcolor[HTML]{ADDCBA}0.406          & \cellcolor[HTML]{EAF6ED}0.107          & \cellcolor[HTML]{D5EDDC}0.207          & \cellcolor[HTML]{DAEFE0}0.186          & \cellcolor[HTML]{A6D9B3}0.442          \\
\textbf{VectorExtrema*}    & \cellcolor[HTML]{ADDCB9}\textbf{0.408} & \cellcolor[HTML]{97D3A7}0.515          & \cellcolor[HTML]{ECF7EF}0.095          & \cellcolor[HTML]{AFDDBB}0.397          & \cellcolor[HTML]{9ED6AD}0.481          & \cellcolor[HTML]{91D0A2}0.543          & \cellcolor[HTML]{7FC993}0.632          & \cellcolor[HTML]{B3DFBE}0.379          & \cellcolor[HTML]{A7DAB5}0.434          & \cellcolor[HTML]{97D3A7}0.517          & \cellcolor[HTML]{84CB97}0.607          \\
\textbf{GreedyMatching*}   & \cellcolor[HTML]{B4DFC0}0.371          & \cellcolor[HTML]{9BD5AB}0.493          & \cellcolor[HTML]{FFFFFF}\red{-0.046}         & \cellcolor[HTML]{C5E6CE}0.29           & \cellcolor[HTML]{9DD5AC}0.484          & \cellcolor[HTML]{A8DAB6}0.43           & \cellcolor[HTML]{80C993}0.629          & \cellcolor[HTML]{C4E6CD}0.295          & \cellcolor[HTML]{A8DAB5}0.433          & \cellcolor[HTML]{ACDCB8}0.413          & \cellcolor[HTML]{85CB98}0.603          \\
\textbf{USE*}              & \cellcolor[HTML]{B8E1C3}0.35           & \cellcolor[HTML]{A4D8B2}0.449          & \cellcolor[HTML]{DBF0E1}0.178          & \cellcolor[HTML]{BAE2C5}0.343          & \cellcolor[HTML]{A3D8B1}0.456          & \cellcolor[HTML]{A2D8B1}0.459          & \cellcolor[HTML]{84CB97}0.609          & \cellcolor[HTML]{BDE3C7}0.326          & \cellcolor[HTML]{ABDBB8}0.417          & \cellcolor[HTML]{A8DAB5}0.433          & \cellcolor[HTML]{89CD9B}0.586          \\
\textbf{WMD}               & \cellcolor[HTML]{B1DEBD}0.389          & \cellcolor[HTML]{8DCE9E}0.565          & \cellcolor[HTML]{E2F3E6}0.146          & \cellcolor[HTML]{ACDCB8}0.413          & \cellcolor[HTML]{97D3A7}0.515          & \cellcolor[HTML]{8ECF9F}0.561          & \cellcolor[HTML]{7BC78F}0.653          & \cellcolor[HTML]{B8E1C3}0.351          & \cellcolor[HTML]{B1DEBD}0.389          & \cellcolor[HTML]{8DCE9E}\textbf{0.564} & \cellcolor[HTML]{77C58C}0.671          \\
\textbf{BertScore*}        & \cellcolor[HTML]{ACDCB8}\textbf{0.413} & \cellcolor[HTML]{7CC790}\textbf{0.646} & \cellcolor[HTML]{BEE3C8}\textbf{0.325} & \cellcolor[HTML]{A0D7AE}\textbf{0.471} & \cellcolor[HTML]{8ECF9F}\textbf{0.56}  & \cellcolor[HTML]{7DC891}\textbf{0.643} & \cellcolor[HTML]{6BC082}\textbf{0.73}  & \cellcolor[HTML]{96D2A6}\textbf{0.522} & \cellcolor[HTML]{90D0A1}0.549          & \cellcolor[HTML]{8ECFA0}0.558          & \cellcolor[HTML]{78C68D}0.666          \\
\textbf{MoverScore*}       & \cellcolor[HTML]{B5DFC0}0.368          & \cellcolor[HTML]{85CB97}0.605          & \cellcolor[HTML]{CFEBD6}0.24           & \cellcolor[HTML]{A7DAB5}0.434          & \cellcolor[HTML]{A0D7AF}0.468          & \cellcolor[HTML]{88CC9B}0.587          & \cellcolor[HTML]{82CA95}0.62           & \cellcolor[HTML]{A4D8B2}0.449          & \cellcolor[HTML]{A2D7B0}0.46           & \cellcolor[HTML]{94D1A4}0.53           & \cellcolor[HTML]{8CCE9D}0.571          \\
\textbf{Stanza+Snomed*}    & \cellcolor[HTML]{B4DFC0}0.37           & \cellcolor[HTML]{A0D7AF}0.468          & \cellcolor[HTML]{EAF6ED}0.106          & \cellcolor[HTML]{B8E1C3}0.35           & \cellcolor[HTML]{A6D9B3}0.443          & \cellcolor[HTML]{94D1A5}0.529          & \cellcolor[HTML]{7FC893}0.633          & \cellcolor[HTML]{A6D9B4}0.441          & \cellcolor[HTML]{99D4A9}0.504          & \cellcolor[HTML]{A4D8B2}0.45           & \cellcolor[HTML]{8ECFA0}0.557

    \end{tabular}
    \caption{Pearson's correlation coefficients for each metric and each human judgement. In bold are the top three scores per column. \wip{All the metrics marked with an asterisk (\textsuperscript{*}) are inversely correlated with the given criterion (e.g. higher post-edit time means worse note, but higher ROUGE score means better note)}; the sign of the coefficient is inverted for ease of visualisation. Coefficients in red are not statistically significant (with p > 0.05).}
    \label{tab:metrics-correlation-pearson}
\end{table*}

\section{Correlation Matrices}
\label{app:correlation-matrices}

\begin{figure*}
    \centering
    \includegraphics[width=1.0\textwidth]{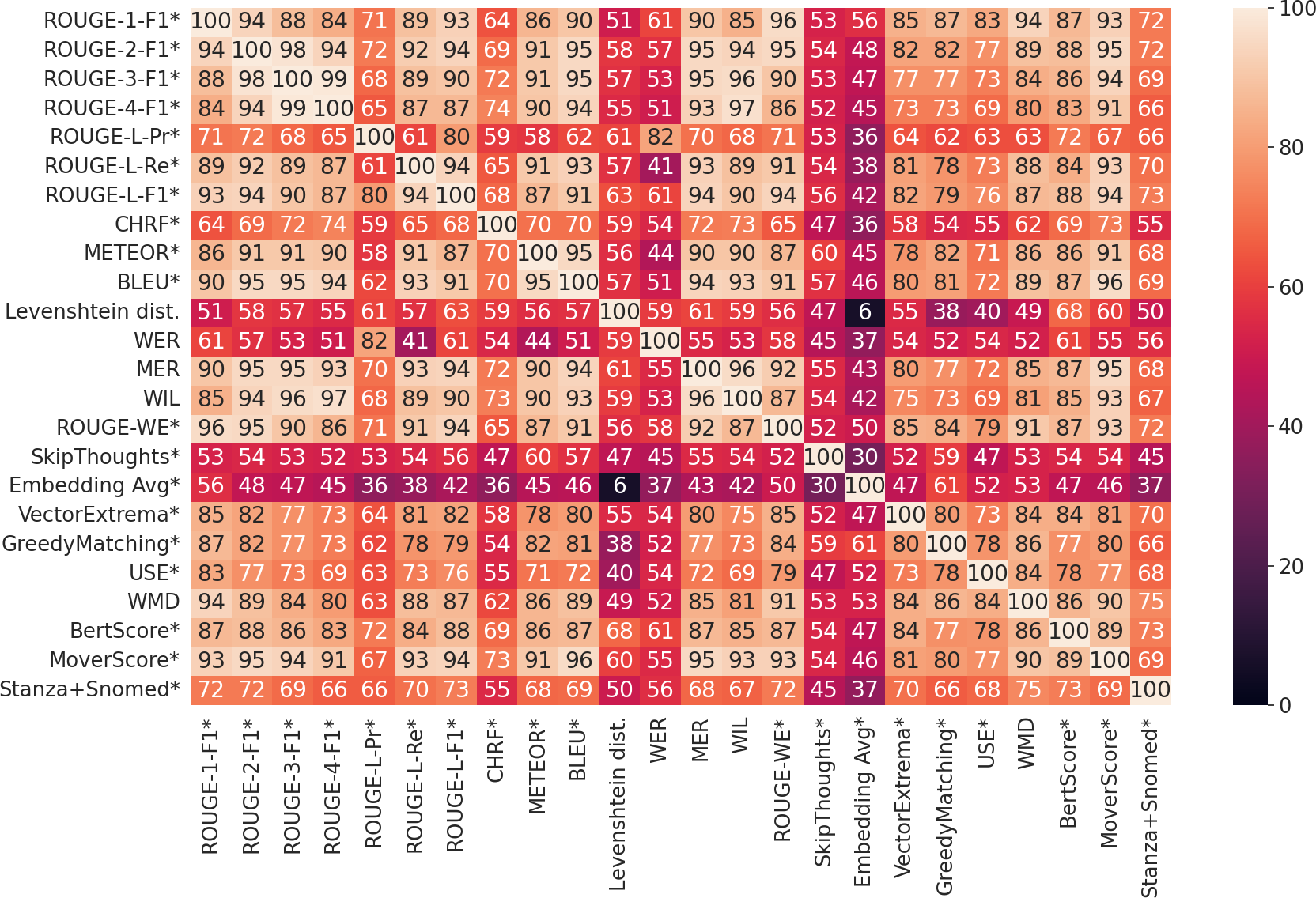}
    \caption{Spearman's correlation matrix between automatic metrics by using all three references and taking the average score. Values represented as percentages for ease of visualisation.}
    \label{fig:correlation-matrix-spearman-avg}
\end{figure*}

\begin{figure*}
    \centering
    \includegraphics[width=1.0\textwidth]{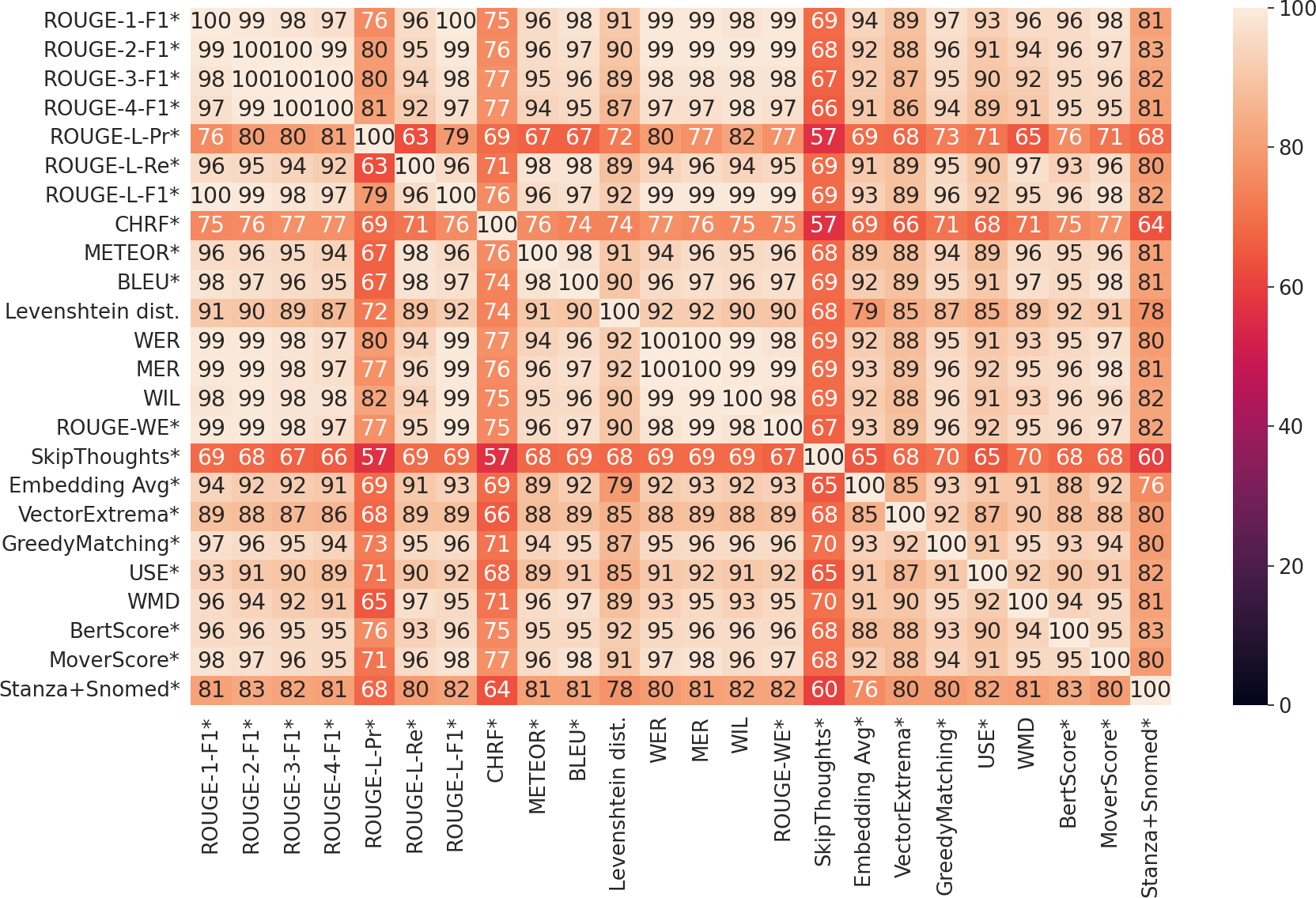}
    \caption{Spearman's correlation matrix between automatic metrics by using all three references and taking the maximum score. Values represented as percentages for ease of visualisation.}
    \label{fig:correlation-matrix-spearman-max}
\end{figure*}

\begin{figure*}
    \centering
    \includegraphics[width=1.0\textwidth]{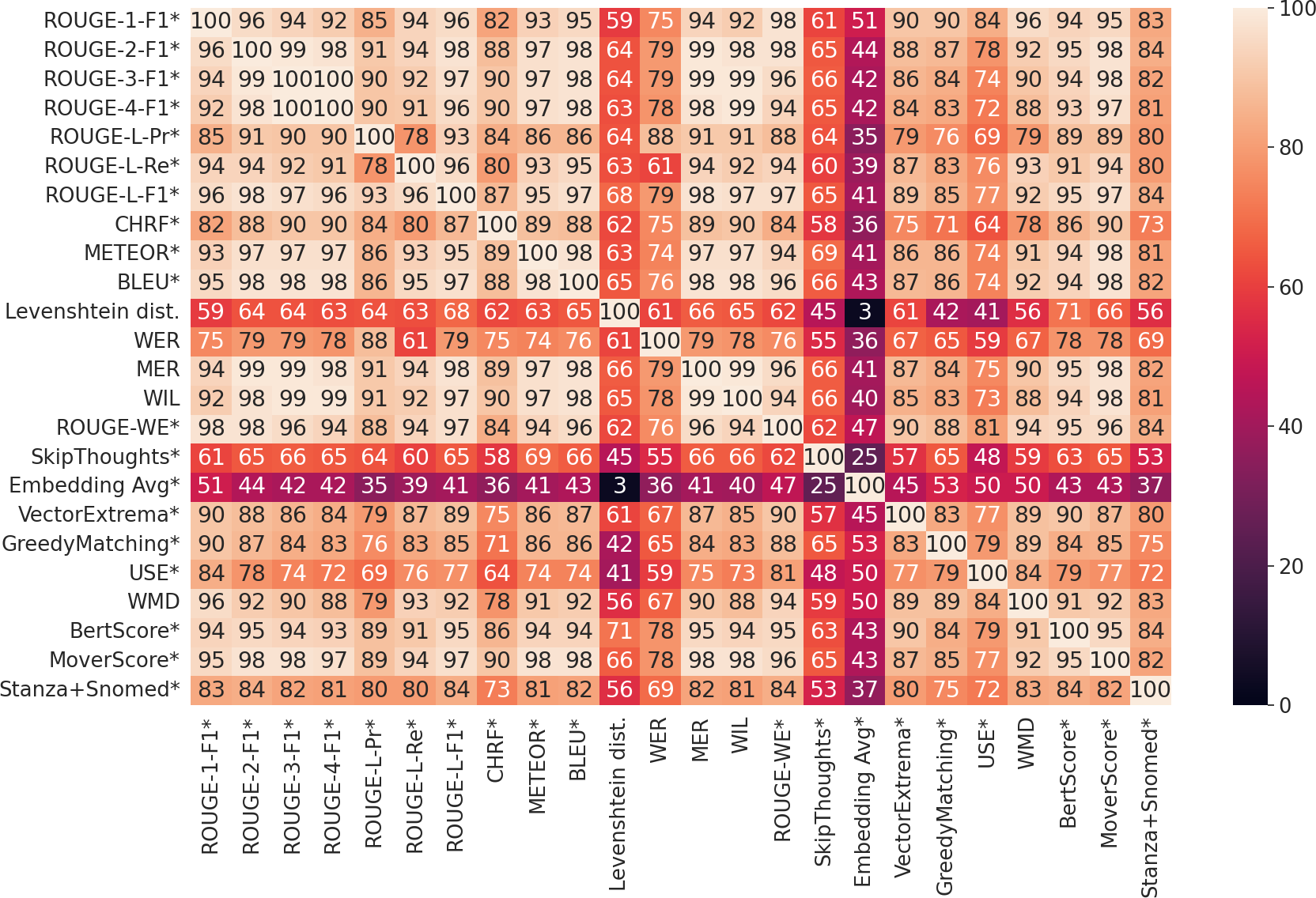}
    \caption{Pearson's correlation matrix between automatic metrics by using all three references and taking the average score. Values represented as percentages for ease of visualisation.}
    \label{fig:correlation-matrix-pearson-avg}
\end{figure*}

\begin{figure*}
    \centering
    \includegraphics[width=1.0\textwidth]{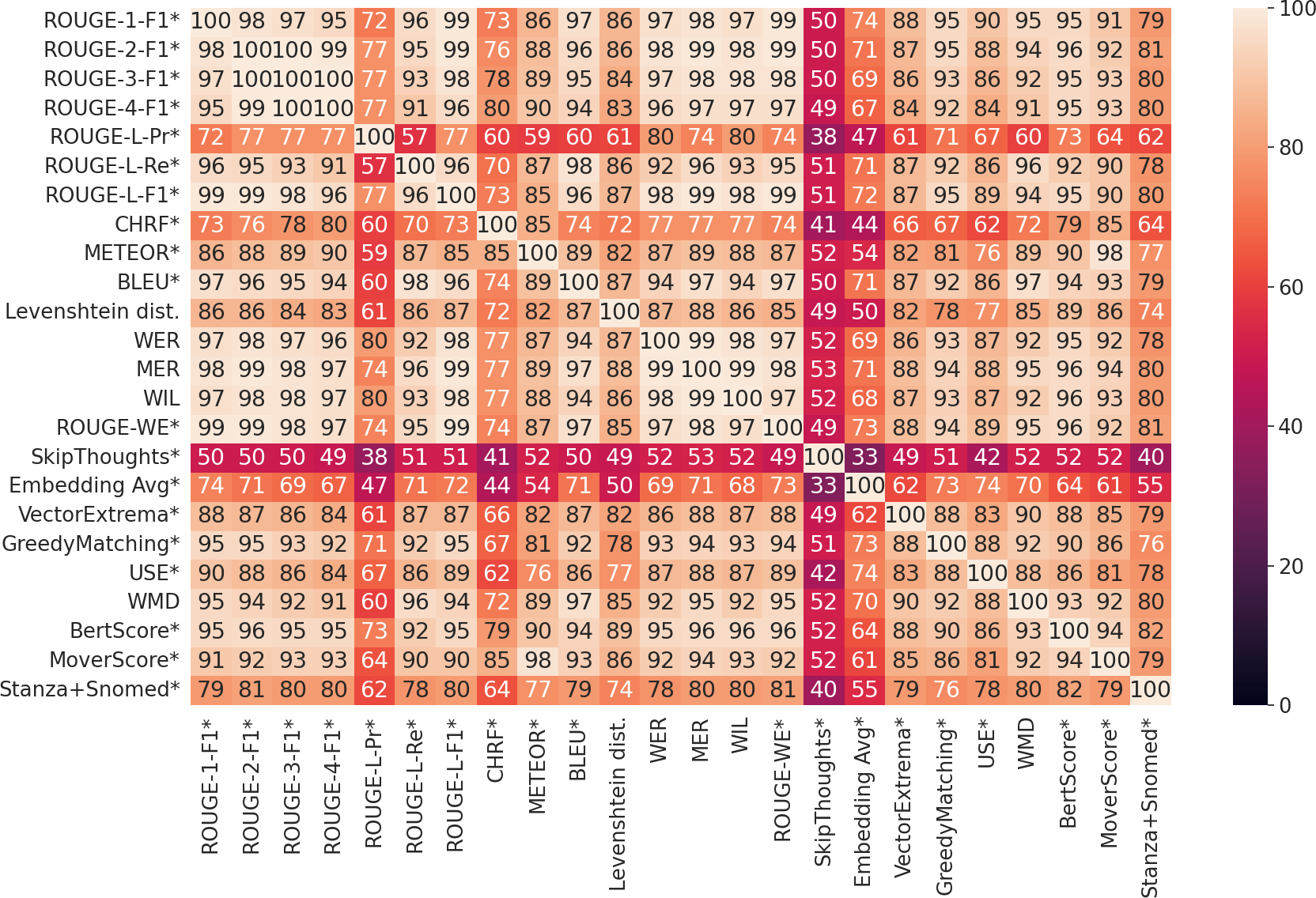}
    \caption{Pearson's correlation matrix between automatic metrics by using all three references and taking the maximum score. Values represented as percentages for ease of visualisation.}
    \label{fig:correlation-matrix-pearson-max}
\end{figure*}

\wip{We compute correlation matrices with Spearman's and Pearson's coefficients for all automatic metrics, considering all three references aggregated by taking the average and maximum scores respectively (Figures \ref{fig:correlation-matrix-spearman-avg}, \ref{fig:correlation-matrix-spearman-max}, \ref{fig:correlation-matrix-pearson-avg}, \ref{fig:correlation-matrix-pearson-max}).}


\end{document}